\documentclass[10pt,twocolumn,letterpaper]{article}

\usepackage{iccv}
\usepackage{times}
\usepackage{epsfig}
\usepackage{graphicx}
\usepackage{amsmath}
\usepackage{amssymb}

\usepackage{pgfplots}
\usepackage{algpseudocode}
\pgfplotsset{compat=1.10}
\usepackage{tikz}

\usepackage{array}
    \newcolumntype{P}[1]{>{\centering\arraybackslash}p{#1}}
    \newcolumntype{M}[1]{>{\centering\arraybackslash}m{#1}}
    
\newcommand{\newloss}{Mismatch Loss} 
    
\definecolor{supurple}{HTML}{674EA7}
\definecolor{forestgreen}{HTML}{6AA84F}
\definecolor{rockbrown}{HTML}{B45F06}
\usepackage{array}
\newcolumntype{L}{>{\centering\arraybackslash}m{3cm}}

\usepackage[pagebackref=true,breaklinks=true,letterpaper=true,colorlinks,bookmarks=false]{hyperref}

\iccvfinalcopy 

\begin{document}

\title{Counting Machine Parts}

\author{Benedict Florance Arockiaraj, Elizabeth Dinella, Ankit Billa, Ajay Anand}

\maketitle

\begin{abstract}
Counting objects in an image is a task applicable across many domains. For instance, crowd counting, inventory counting, and cell counting have been the focus of recent research. The major challenges in estimating the count of objects include  overlapping objects, object scale issues, occlusions, and varying lighting conditions. In this report, we explore the problem of counting machine washer parts. Our technique is an extension of FamNet~\cite{Ranjan2021LearningTC} with an additional loss component, trained on the given dataset. We compare to three baseline methods: a traditional image processing pipeline, instance segmentation, and density map estimation. We evaluate the performance of these algorithms by computing the Mean Absolute Error (MAE) and the Root Mean Squared Error (RMSE) between the true object counts and the model outputs. Our approach achieves a performance of \textbf{1.96 MAE}. The video presentation can be found \href{https://drive.google.com/file/d/1QaI-QAzHdPyYpGDtnGV_qW4KoSf-Qgr4/view?usp=sharing}{here}. The code can be found \href{https://github.com/benedictflorance/cis-680-final-project}{here}.
   
\end{abstract}

\section{Introduction}
\noindent The task of object counting has a wide range of applications such as crowd estimation, inventory management, and other industrial processes. Each domain suffers from its own unique challenges, but densely packed objects pose a problem to many. In this report, we study the task of counting densely packed objects. 

Given that object detection systems have made great strides in recent years, they may seem to be a natural solution to object counting. However, our experiments in Section \ref{sec:eval} find the current state of the art object detection systems to struggle with counting densely packed objects. 

To resolve these issues, state of the art techniques have been developed for the task of counting objects in dense environments, focusing on crowd estimation with the goal of increasing crowd safety. In our work, we utilize these algorithms for counting machine parts given their baseline performance in low-light environments with many overlapping objects. 

Our approach builds on FamNet \cite{Ranjan2021LearningTC}, a regression based approach which learns a density map for each image. FamNet uses a Mean Squared Error loss function between the true and predicted density maps. However, this loss function does not take into the account the localized density map error, and only considers errors in the total. In order to solve this problem, we propose an alternative loss function called \textit{mismatch} loss. Since our dataset includes 9 angles for each set of objects, we take the average of predicted counts for each angle, as a post processing step.

We evaluate our approach against several traditional and deep-learning based baselines. We also include experiments on out of distribution data sets to evaluate our techniques generalizability.


\section{Related Work}


\begin{figure*}[!h]
\centering
    {\includegraphics[scale=0.2]{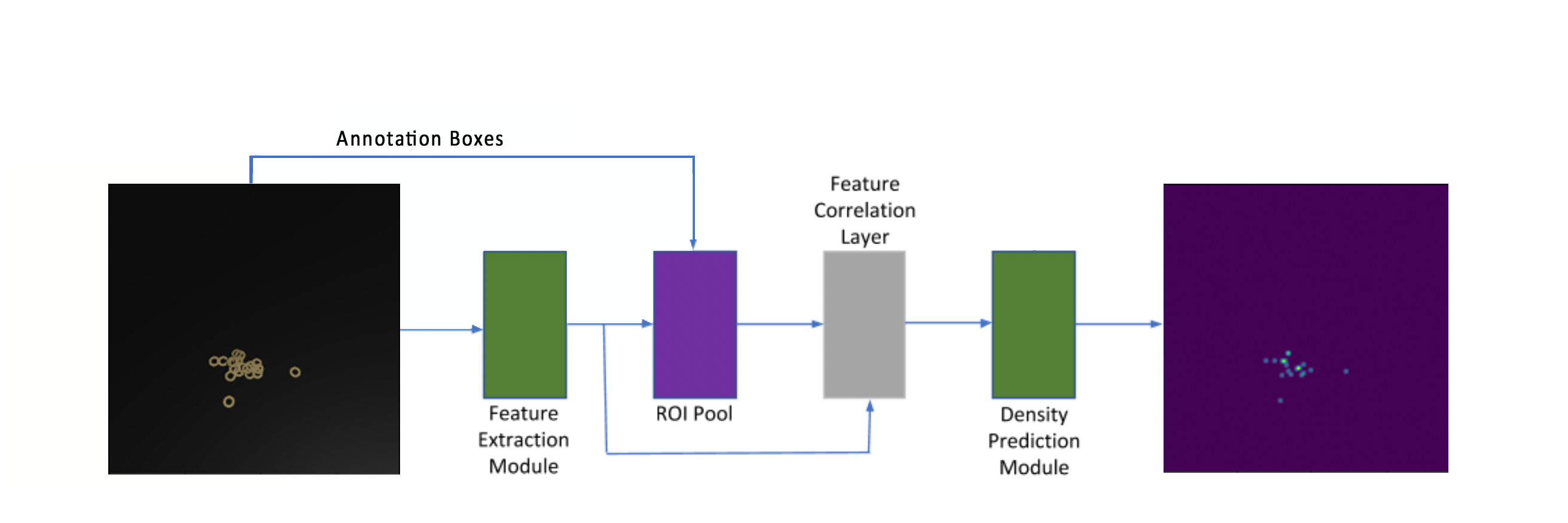}}
\caption{Proposed Model Architecture based on FamNet \cite{Ranjan2021LearningTC}}
\label{arch}
\end{figure*}


\subsection{Traditional Approaches}

\noindent Traditional methods for machine vision operations such as product counting, error control and dimension measurement mainly revolve around image filtering techniques to obtain object counts or even detect objects. One such pipeline was proposed by Baygin \etal \cite{imageproc}, which combines multiple image filtering techniques with thresholding methods to obtain edge maps of each circular object, and then counts the number of circles generated using a Hough Transform.

Since the given dataset specifically consists of circular machine parts, we implement the same pipeline to get baseline results.

\subsection{Object Detection}

There has been much work in object detection~\cite{SINDAGI20183,9156690}, in which the goal is to identify and locate objects in an image. Object detection approaches employ a sliding window to localize objects within a small local region. Initial naive approaches to object counting simply run an object detection model and count the number of detected objects. This naive approach is somewhat effective in low density scenes, but has trouble generalizing to images with a high density of objects. Furthermore, architectures that employ sliding windows are heavyweight and computationally expensive. 

Advances in using whole image detection techniques remedy some of the drawbacks of sliding window based techniques ~\cite{SINDAGI20183,Yang_2020_CVPR} but still have the same drawbacks as object detection algorithms for detecting objects in dense or obscured images. 

Regression based approaches ~\cite{inproceedings, 8490884, Yang_2020_CVPR} address the poor performance of object detection systems to count objects in densely packed scenes. These methods attempt to learn the count directly without locating and counting each object. They directly learn a map between image features and a number of objects. These image features may be local or whole image.

\subsection{Density Estimation}

Recent contributions in the field of object counting have focused on the task of density estimation ~\cite{Lempitsky10b, Arteta14}. These approaches incorporate spatial information and also side-step the difficult task of learning to detect and localize each individual object. However, these techniques are more well suited for large crowds, where the exact count is not needed. For our setting of machine part counting, the number of objects is much less and thus the margin of error should be lower. 

We see a shift in the overall algorithmic approach from traditional to CNN-based methods for object counting. Specifically, recent works use multi-scale features to tackle the scale variation problem \cite{Yang_2020_CVPR}, attention based methods to focus on useful information \cite{9156690} and use auxiliary tasks \cite{Wan_2021_CVPR} to improve counting performance.

\section{Dataset Analysis}
The dataset consists of 8100 images of washer parts in a bundle, with 900 unique images and each image being captured from 9 different angles. The images are of dimensions $1080\times1080$, with the annotations file having bounding boxes and masks for every image in COCO format. Processing the \texttt{annotation.json} file in the dataset shows that the average object count per image is 25.969 and the median object count per image is 21.

\section{Proposed Method} 
We base our proposed method off of \cite{Ranjan2021LearningTC}, which uses a few-shot regression based model to perform object counting. They use the image, along with a few exemplar object boxes from the image, to predict a density map, that estimates the count of objects in the image.

\subsection{Ground Truth Density Construction}
To generate the density map, we first use Gaussian smoothing with a dynamic window size dependent on the dataset. Deviating from \cite{Ranjan2021LearningTC}, where the dataset is specifically annotated point-wise, we instead fix the point annotations to the center-of-mass of the object polygon. Having computed the point annotations, we calculate the average distance between every point and its 1-nearest neighbor.  The computed average distance acts as the dynamic window size for the Gaussian smoothing. We use the same hyperparameters for smoothing (std dev $\sigma = window\_size/4$) as mentioned in \cite{Ranjan2021LearningTC}.

\subsection{Network Architecture}
We use the FamNet architecture proposed in \cite{Ranjan2021LearningTC}, with slight adjustments to the overall structure. Figure \ref{arch} shows the pipeline of the network. The network consists of two components: a multi-scale feature extraction module and a density prediction module.  The feature extraction component uses pre-trained ImageNet ResNet-50~\cite{He2015} backbone for feature extraction. We make a number of modifications to FamNet for machine part counting. 

FamNet, is unique in that it can adapt to counting any particular class at test time, regardless of the class label. Given only a few exemplar bounding boxes, the model can adapt to counting objects of that class. As its title, \textit{Learning to Count Everything} suggests, FamNet is highly generalizable to different domains. 

In our setting, we are not concerned with adapting to different classes at test time. At training and inference time, we have a single class (washer) which we aim to count. Rather than give exemplar bounding boxes of washers, as in the original paper, we give all bounding boxes around each washer. In order to accommodate for this, we obtain multi-scale features of the bounding boxes by performing ROI pooling on the convolutional feature maps. The convolutional feature maps after the fourth block of the backbone are passed to the density estimation module. 

To make the density prediction module category agnostic, the authors take a correlation between the pooled feature map from the objects and the image feature extraction map. The density estimation module has a series of five convolution blocks and three upsampling layers interleaved. The 2D density map is obtained using a final 1$\times$1 convolution layer. By summing the density map, we can obtain the floating-point count of the number of objects in the input image.

\subsection{Mismatch Loss}
The original FamNet~\cite{Ranjan2021LearningTC} architecture was trained with a single loss function: 
\begin{equation}
\label{eq:mse}
    density\_MSE\_loss = MSE(pred\_map, gt\_map)
\end{equation}

The \texttt{density\_MSE\_loss} computes the mean squared error between the predicted and ground truth density maps.  

We propose an additional loss function: \texttt{Mismatch Loss}. Intuitively, the \newloss penalizes pixels in the predicted density map which do not actually correspond to objects in the original image. As ground truth, we generate masks where a value of 1 indicates that the pixel doesn't have any objects of interest and a value of 0 indicates otherwise. The network should ideally predict a density map with 0 values in places where the mask has a value of 1. We penalize when the model predicts objects where a mask is not covering. To create masks, we convert the mask polygons from the annotation to grayscale masks.

Formally,  
\begin{equation}
\label{eq:mismatch}
    mismatch\_loss = \Sigma_{i,j} (I(i,j))
\end{equation}
\begin{equation}
    I(i,j) = pred\_map[i][j] == 1 \land mask[i][j] == 1
\end{equation}
$I$ is an indicator function which returns 1 if the \texttt{pred\_map} and \texttt{mask} both have a value of 1 at the location \texttt{i,j}.  
If \texttt{pred\_map} is equal to 1, this means the model predicted an object in that location. On the other hand, if \texttt{mask} is equal to 1, this means that the mask is white, and does not include an object. Thus, they are mismatching and should be penalized. 

The \textit{mismatch\_loss} is a sum over pixels and can thus range up to the number of pixels in the entire image. To scale this appropriately, we add a hyperparameter $\lambda$ to balance both the losses. We use a $\lambda$ of $10^{-9}$ to scale down the mismatch loss.

Lastly, we combine the loss functions in equations \ref{eq:mse}, \ref{eq:mismatch} to train FamNet with a single unified loss.

\begin{equation}
    loss = density\_MSE\_loss + \lambda * mismatch\_loss
\end{equation}

\subsection{Angle Aggregation}
\noindent The provided washer dataset includes multiple perspectives for each scene. As a post processing step, we perform aggregation of \textit{votes} from the model's inferred count at each angle. We experimented with three different aggregation methods: \texttt{max}, \texttt{min}, and \texttt{average}. We empirically found that \texttt{average} performed the best and improved performance over the model without aggregation. 

\begin{figure*}[h]
\centering
\setlength\tabcolsep{1.5pt}
\begin{tabular}{c c c}
 RGB & Image Processing & Mask-RCNN \\[1ex]

{\includegraphics[width=1.5in,height=1.5in,angle=0]{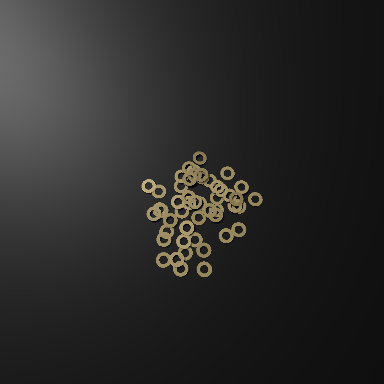}}
&
{\includegraphics[width=1.5in,height=1.5in,angle=0]{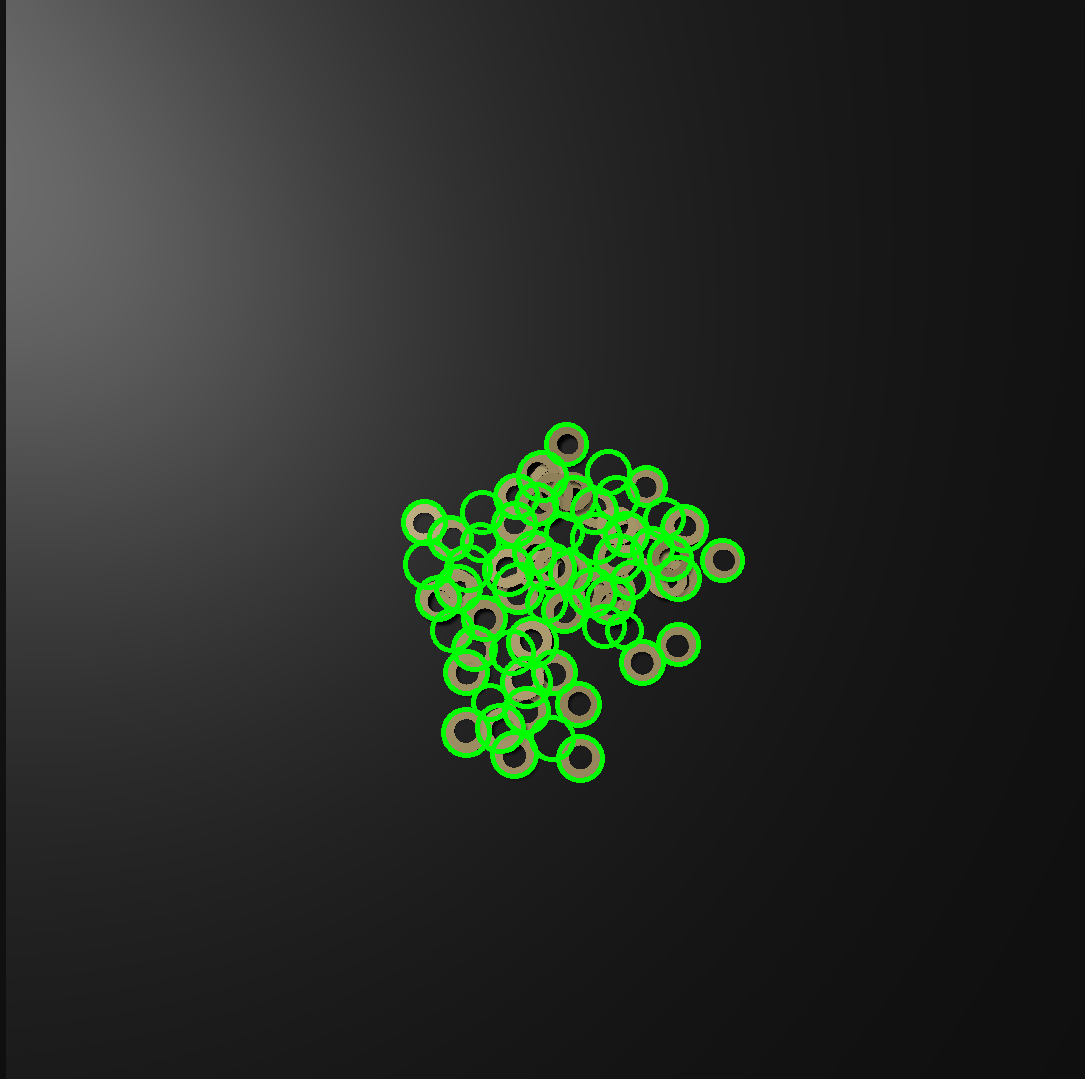}}
&
{\includegraphics[width=1.5in,height=1.5in,angle=0]{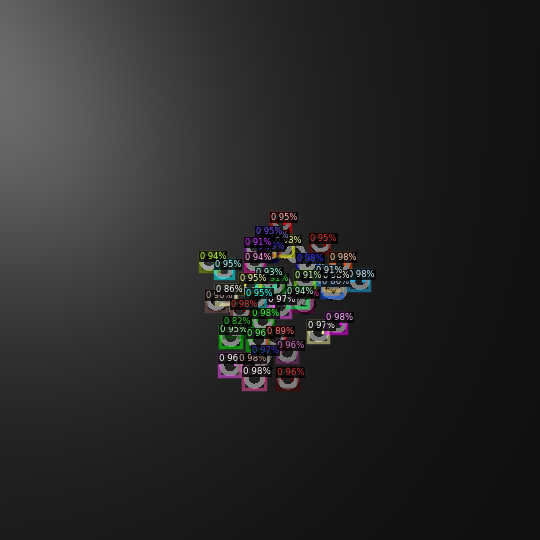}}\\[1ex]

\end{tabular}
\caption{Baseline Predictions}
\label{baseline}
\end{figure*}


\begin{figure*}[h]
\centering
\setlength\tabcolsep{1.5pt}
\begin{tabular}{c c c}
 RGB & Ground Truth Density & Predicted Density \\[1ex]

{\includegraphics[width=1.5in,height=1.5in,angle=0]{images/134_angle4_img_rgb.png}}
&
{\includegraphics[width=1.5in,height=1.5in,angle=0]{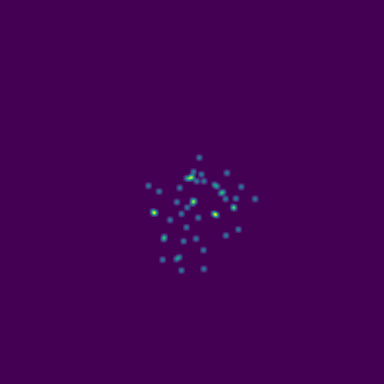}}
&
{\includegraphics[width=1.5in,height=1.5in,angle=0]{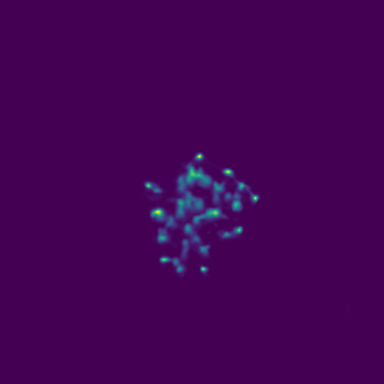}}\\

{\includegraphics[width=1.5in,height=1.5in,angle=0]{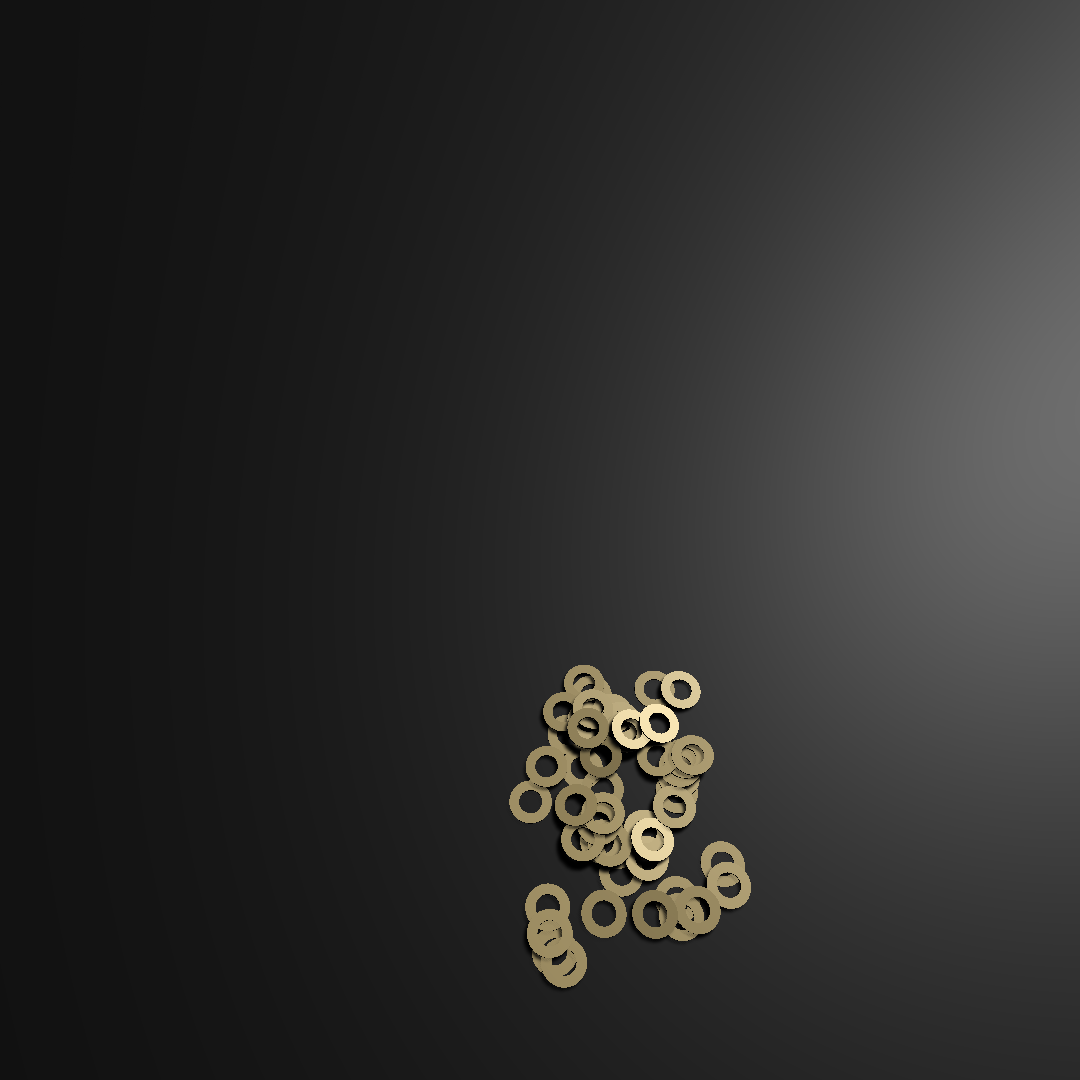}}
&
{\includegraphics[width=1.5in,height=1.5in,angle=0]{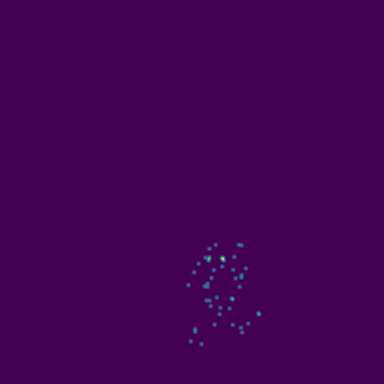}}
&
{\includegraphics[width=1.5in,height=1.5in,angle=0]{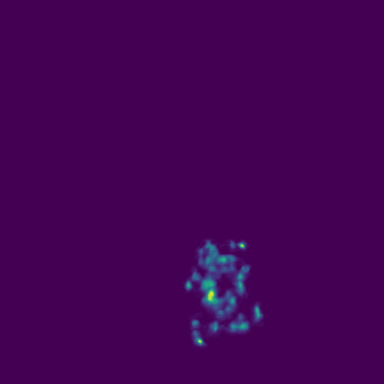}}\\[1ex]

\end{tabular}
\caption{Predictions of the proposed method}
\label{famnet_preds}
\end{figure*}


\section{Experiments}
\noindent For the purpose of our project, we evaluate two baseline methods, with the first being the traditional image processing pipeline proposed in \cite{imageproc}. The second baseline is an object instance segmentation network (MaskRCNN) used to  count the number of machined parts in the provided data set. 

Drawing inspiration from \cite{Lempitsky10b} and some of the recent works in this domain \cite{inproceedings, 9156690, Ranjan2021LearningTC, Wan_2021_CVPR}, we planned to improve the performance on the given task of machined-part counting over pre-existing networks which are generally tailored towards crowd counting tasks.

Our final technique is primarily based on the work done on learning to count objects of diverse visual categories using a few shot learning model \cite{Ranjan2021LearningTC}. We adapted the model to our machine part task and added a new postprocessing step. We also experimented with an additional loss function.


\begin{figure*}[h]
\centering
\setlength\tabcolsep{1.5pt}
\begin{tabular}{c c c c}
 RGB & Otsu Level 2 & Otsu Level 4 & Otsu Level 6\\[1ex]

{\includegraphics[width=1.5in,height=1.5in,angle=0]{images/134_angle4_img_rgb.png}}
&
{\includegraphics[width=1.5in,height=1.5in,angle=0]{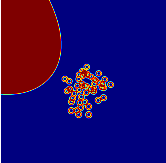}}
&
{\includegraphics[width=1.5in,height=1.5in,angle=0]{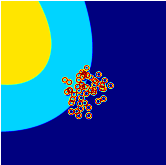}}
&
{\includegraphics[width=1.5in,height=1.5in,angle=0]{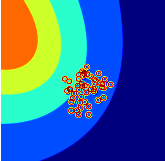}}\\[1ex]

\end{tabular}
\caption{Multi-level Otsu Thresholding results}
\label{otsu}
\end{figure*}


\begin{figure*}[h]
\centering
\setlength\tabcolsep{1.5pt}
\begin{tabular}{c c c c}
 RGB & DPT-Hybrid & DPT-Large & DPT-Hybrid-NYUV2\\[1ex]

{\includegraphics[width=1.5in,height=1.5in,angle=0]{images/134_angle4_img_rgb.png}}
&
{\includegraphics[width=1.5in,height=1.5in,angle=0]{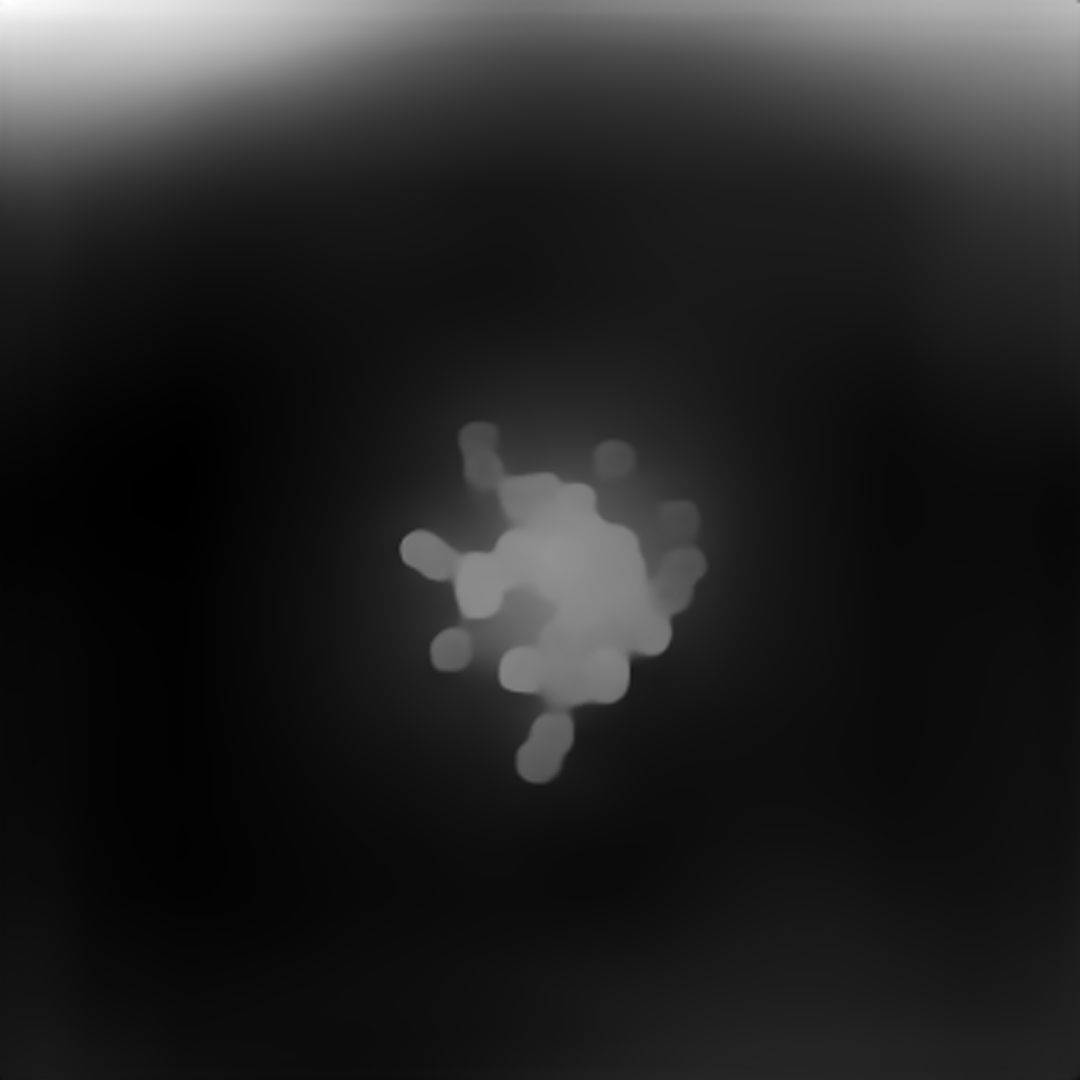}}
&
{\includegraphics[width=1.5in,height=1.5in,angle=0]{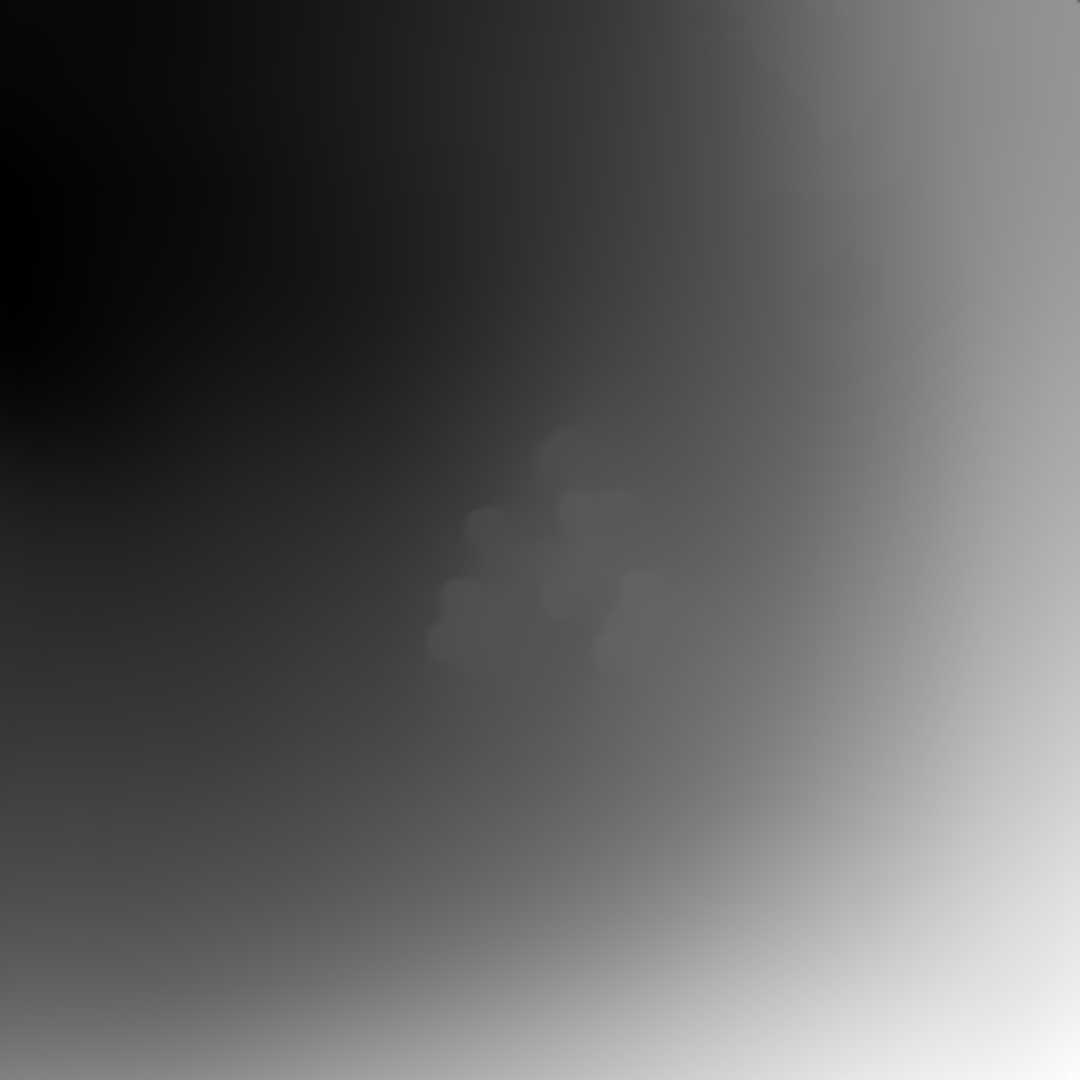}}
&
{\includegraphics[width=1.5in,height=1.5in,angle=0]{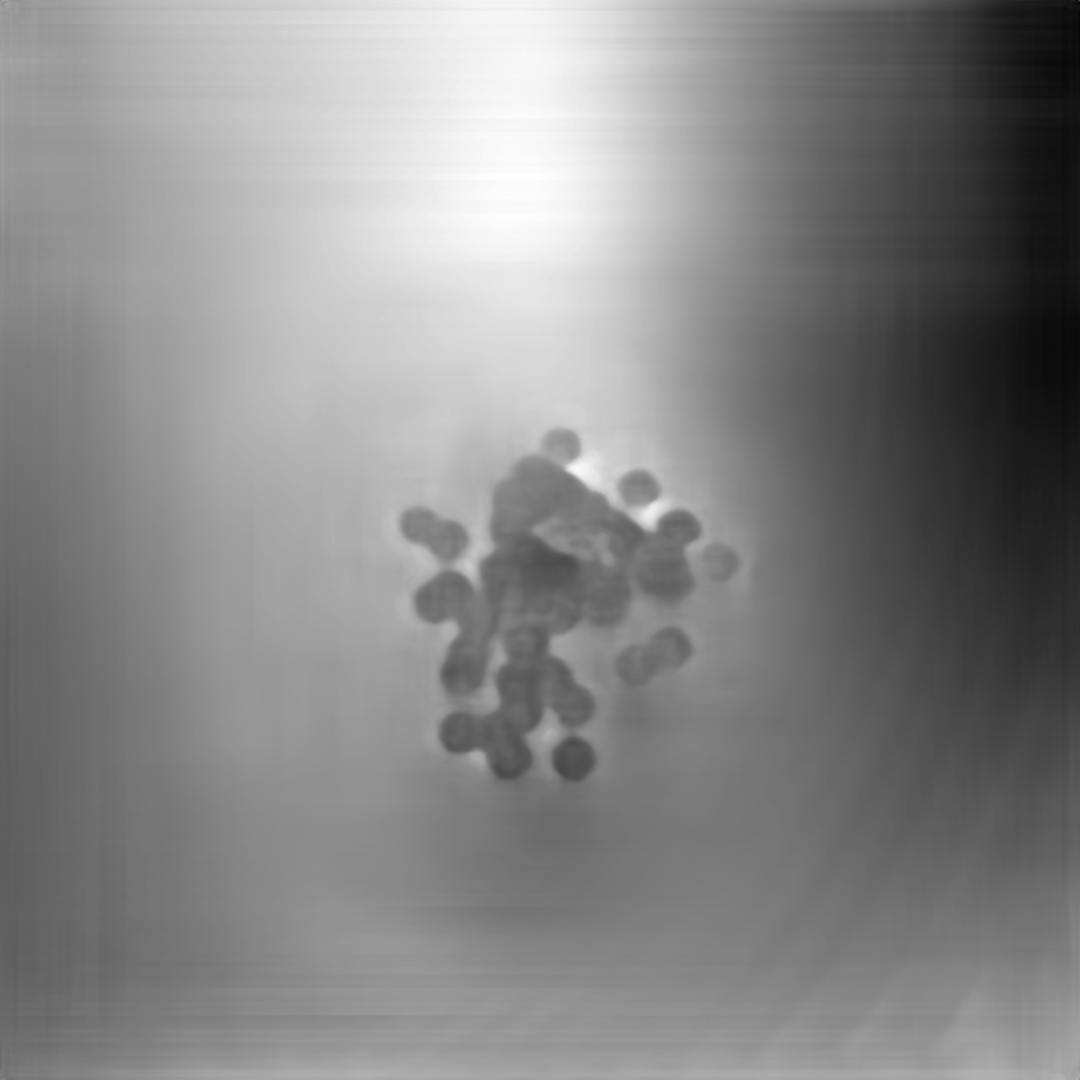}}\\[1ex]

\end{tabular}
\caption{Depth Maps generated by DPT \cite{Ranftl2021}}
\label{dpt}
\end{figure*}


\subsection{Image Processing Pipeline}
One of the baseline techniques implemented was to use traditional image processing techniques to obtain an estimate of the number of regularly shaped objects present in the provided data set. Given that our industry challenge task consisted of counting regularly shaped circular objects, we were able to obtain relatively decent results using such techniques similar to the approach devised in \cite{imageproc}. The proposed method consisted of splitting the data set images to component channels. Using the saturation channel provided the best results for object counting. A Gaussian blur is now applied to the resulting grey scale image to remove noise. The de-noised image is then Otsu thresholded to separate the foreground from the background. A Sobel filter is then applied to the foreground image to perform edge detection. Hough circle detection is then performed on the edge detected image to obtain a count of the number of circular objects as depicted in Figure \ref{baseline}. We also tuned the parameters of the Hough circles algorithm to obtain better detection performance.

\subsection{Instance Segmentation (Mask-RCNN)}
The first is an out of the box object detection technique, Detectron2~\cite{wu2019detectron2}. Object detection methods use a sliding window to identify objects locally.  As a naive baseline we sum the number of object identifications in each local window to achieve a global object count. In particular we compare to the Detectron2 model: \texttt{mask\_rcnn\_R\_50\_FPN\_3x}. That is, a MaskRCNN~\cite{mask-rcnn} with a ResNet-50~\cite{He2015} and a three level feature pyramid network~\cite{fpn} backbone.  

The original Detectron2 model was pretrained on the COCO 2017-train dataset. We include evaluation against the pretrained model as well as a finetuned model trained on the washer dataset. The performance improved from an MAE of 26.1963 to 3.7889 after finetuning. This is expected as the COCO 2017 dataset is drastically out of distribution containing large objects such as people and animals with less clutter than the washer dataset. Model predictions for the machine parts using Mask-RCNN are illustrated in Figure \ref{baseline}.

\subsection{FamNet}
To train FamNet, we minimize the proposed loss function that has density and mismatch components. We use Adam optimizer, with hyperparameters as mentioned in \cite{Ranjan2021LearningTC} (learning rate of $10^{-5}$ and batch size of 1). Every image is resized to 384x384 from the standard 1080x1080 image dimension in the washer dataset. We use a train-dev-test split of 80:10:10, and split images in such a way that all angles of a particular washer setup image goes in the same bucket. This is done so as to avoid any memorization of test set images that can potentially occur in a random split. We train for a total of 20 epochs for each of the experiments.

\subsection{Depth-based Image Separation}
Since our FamNet model already gave a very close count for input images with just a tiny MAE of 2, we noticed that the major performance drop was because of occlusions. To tackle this, we wanted to perform depth separation for our images. Our motive for the following experiments was to divide the input images into multiple depths. Once we have the depth separated images, we planned to run the FamNet model on each of the depth images individually, and take a net sum of the predictions at each level. 

\subsubsection{Multi-Level Otsu Thresholding}
Multi-level Otsu thresholding was carried out to improve performance and possibly extract features from the given dataset, as opposed to the conventional adaptive Otsu thresholding. Multi-Otsu calculates several thresholds, determined by a provided count of the desired classes. The default number of classes is 3: for obtaining three classes, the algorithm returns two threshold values. By varying the number of required thresholds, Multi-Otsu thresholding can be used to segment objects in an image into different depth classes based on their grey intensity levels.


\begin{table*}[t]
\centering 
\begin{tabular}{|l |c | c|} 
\hline 
Method & mae↓ & rmse↓\\
\hline 
Image Processing \cite{imageproc} & 29.0086 & 37.06189\\
Mask-RCNN \cite{wu2019detectron2} (without fine-tuning) &  26.1963 & 27.9758\\
Mask-RCNN \cite{wu2019detectron2} (with fine-tuning) & 3.7889 & 5.0149 \\
\hline 
\end{tabular}
\caption{Baseline Performance}
\label{baseline_perf}
\end{table*}

\begin{table*}[t]
\centering 
\begin{tabular}{|l |c | c|} 
\hline 
Method & mae↓ & rmse↓\\
\hline 
FamNet \cite{Ranjan2021LearningTC} & 2.60 &  3.64 \\
FamNet \cite{Ranjan2021LearningTC} + Angle Aggregation (Min) & 2.78 & 4.07 \\
FamNet \cite{Ranjan2021LearningTC} + Angle Aggregation (Max) & 3.26 & 3.74 \\
FamNet \cite{Ranjan2021LearningTC} + Angle Aggregation (Mean) & \textbf{1.96} & \textbf{2.70} \\
FamNet \cite{Ranjan2021LearningTC} + \newloss &  2.97 & 3.91 \\
FamNet \cite{Ranjan2021LearningTC} + \newloss + Angle Aggregation (Min) &  3.06 &   4.35 \\
FamNet \cite{Ranjan2021LearningTC} + \newloss + Angle Aggregation (Max) &  3.95 &  4.74 \\
FamNet \cite{Ranjan2021LearningTC} + \newloss + Angle Aggregation (Mean) & 2.69  &  3.52 \\[1ex]
\hline 
\end{tabular}
\caption{Performance of the proposed model}
\label{proposed_perf}
\end{table*}


\subsubsection{Dense Prediction Transformer}
We implement the DPT model \cite{Ranftl2021} for our problem to obtain depth maps for each image of the dataset, given its recent success. The purpose of this experiment was to separate machine parts in an image based on their depth i.e. the intensity of pixels in the depth map, so as to get a count of occluded objects better. We primarily used 3 variants of DPT: DPT-Hybrid, DPT-Large \& DPT-Hybrid Finetuned on the NYU-Depth-V2 dataset. Results of this experiment are illustrated in Figure \ref{dpt}.

\section{Evaluation and Analysis}
\label{sec:eval}
\subsection{Evaluation}
\noindent We evaluate the performance of the models on the test set using the following error metrics:
\begin{itemize}
\item {\bf Mean Absolute Error }(MAE): The MAE represents the average absolute difference between predicted and ground-truth values of the number of objects detected in an image.
\item {\bf Root Mean Squared Error }(RMSE): The RMSE represents the square root of the second sample moment of the differences between predicted values and observed values or the quadratic mean of these differences.
\end{itemize}

\noindent Quantitative results of the baseline methods are illustrated in Table \ref{baseline_perf}, and the performance of our proposed method is illustrated in Table \ref{proposed_perf}.

\subsection{Analysis}
\noindent In performing experiments with image processing techniques, it was found that it was difficult to fine tune the parameters of the Hough circles function to accurately detect all the objects while avoiding false detections due to noise in the image.
While implementing Multi-Otsu thresholding, it was found that due to the nature of the provided data set, objects could not be wholly separated into different depth levels via their adaptive grey scale intensity.

A similar result was observed for the DPT model \cite{Ranftl2021} while generating the depth maps, with the DPT-Large and DPT-Hybrid models not giving accurate enough depth separations. This led to duplicate counts of objects from different thresholding levels leading to large deviations from ground truth values. Changes in lighting conditions also affected the performance of multi level Otsu thresholding with brighter areas appearing in the foreground even if they are supposed to be part of the background as shown in Figure \ref{otsu}. Since the DPT-Hybrid model fine-tuned on NYU-Depth-V2 \cite{Silberman:ECCV12} is more accustomed to indoor scenes, it gives reasonable depth maps. However, due to compute and time constraints, we could not utilize these depth maps well. These results disabused us of the notion that using these techniques would improve network performance and they were relegated to being unsuccessful experiments.

FamNet~\cite{Ranjan2021LearningTC} fine tuned on washers, achieves an MAE less than 2. In comparison to the Mask-RCNN and image processing techniques (3.79 and 29.01 respectively), FamNet performs well. 

Our post processing aggregation technique reduced MAE from 2.60 to 1.96 with averaging. We also found that \texttt{min} and \texttt{max} aggregation increased the MAE. 

Through our empirical results in Table~\ref{proposed_perf}, we observed that the \texttt{min} aggregation has better performance than the \texttt{max} aggregation, suggesting that our technique may be over-estimating. 
To combat the over-estimation, we added the mismatch loss to penalize predictions in the density map where there is no object of interest in the ground truth. As one can observe from \ref{proposed_perf}, the experiment with the mismatch loss increased the MAE instead. Through a closer post-mortem of the experiments, we identify that although the root cause of the minor errors were due to over counting of objects, the overestimate happens at occluded regions where there's already an object prediction. This way, the mismatch loss did not help much to fix the errors. However, we do believe that the addition of mismatch loss could help when we try to count objects of diverse scales and visual categories, unlike the ones in the washer parts dataset that has a plain black background.

\section{Conclusion}
We find that it is definitely possible to perform the task of counting machined washer parts using deep learning networks with a high degree of accuracy. The robustness of such a model to generalize to other tasks however, is called into question due to the specific nature of our training data set. Given a most general training data set with diverse visual categories, we believe that it would be possible to use this network to perform all manner of dense object counting tasks. Some avenues of future work include utilizing the depth-wise separation techniques mentioned in Section 5.4 in a more effective way or trying to exploit the multi-view nature of the dataset to generate 3D representations of scenes using models like NeRF \cite{mildenhall2020nerf} to obtain more accurate results.


\newpage
\newpage
{\small
\bibliographystyle{ieee_fullname}
\bibliography{bibfile}
}

\end{document}